\documentclass{IEEEtran}
\usepackage{cite}
\usepackage{amsmath,amssymb,amsfonts}
\usepackage{algorithmic}
\usepackage{graphicx}
\usepackage{textcomp}
\usepackage{cite}
\usepackage{amsmath,amssymb,amsfonts}
\usepackage{algorithmic}
\usepackage[ruled]{algorithm2e}
\usepackage{graphicx}
\usepackage{textcomp}
\usepackage{amsmath,amssymb,amsfonts}
\usepackage{algorithmic}
\usepackage{graphicx}
\usepackage{textcomp}
\usepackage{multirow}
\usepackage{stfloats}
\usepackage{multicol}
\usepackage{float}
\usepackage{amsmath,lipsum}
\usepackage[misc,geometry]{ifsym}
\def\BibTeX{{\rm B\kern-.05em{\sc i\kern-.025em b}\kern-.08em
    T\kern-.1667em\lower.7ex\hbox{E}\kern-.125emX}}
\begin{document}
\title{PMP-Swin: Multi-Scale Patch Message Passing Swin Transformer for Retinal Disease Classification}
\author{Zhihan Yang, Zhiming Cheng, Tengjin Weng, Shucheng He, Yaqi Wang, Xin Ye, Shuai Wang
\thanks{This work was supported in part by the Zhejiang Provincial Natural Science Foundation of China under Grant LDT23F01015F01, in part by the National Natural Science Foundation of China under Grant 62201323, and in part by the Natural Science Foundation of Jiangsu Province under Grant BK20220266. \textit{(First author: Zhihan Yang.) (Corresponding authors: Yaqi Wang; Xin Ye; Shuai Wang.)}}
\thanks{Zhihan Yang is with School of Mechanical, Electrical and Information Engineering, Shandong University, Weihai, China (e-mail: yangzhihan@mail.sdu.edu.cn).}
\thanks{Zhiming Cheng is with School of Automation, Hangzhou Dianzi University, Hangzhou, China (email: czming@hdu.edu.cn).}
\thanks{Tengjin Weng is with School of Computer Science and Technology, Zhejiang Sci-Tech University, Hangzhou, China (email: wtjdsb@gmail.com).}
\thanks{Yaqi Wang is with the College of Media Engineering, Communication University of Zhejiang, Hangzhou, China (e-mail: wangyaqi@cuz.edu.cn).}
\thanks{Shucheng He and Xin Ye are with the Center for Rehabilitation Medicine, Department of Ophthalmology, Zhejiang Provincial People’s Hospital (Affiliated People’s Hospital, Hangzhou Medical College), Hangzhou, China (email: hscflea@163.com; yexinsarah@163.com).}
\thanks{Shuai Wang is with School of Cyberspace, Hangzhou Dianzi University, Hangzhou, China, and Suzhou Research Institute of Shandong University, Suzhou, China (e-mail: shuaiwang.tai@gmail.com).}}
\maketitle

\begin{abstract}
Retinal disease is one of the primary causes of visual impairment, and early diagnosis is essential for preventing further deterioration. Nowadays, many works have explored Transformers for diagnosing diseases due to their strong visual representation capabilities. However, retinal diseases exhibit milder forms and often present with overlapping signs, which pose great difficulties for accurate multi-class classification. Therefore, we propose a new framework named Multi-Scale Patch Message Passing Swin Transformer for multi-class retinal disease classification. Specifically, we design a Patch Message Passing (PMP) module based on the Message Passing mechanism to establish global interaction for pathological semantic features and to exploit the subtle differences further between different diseases. Moreover, considering the various scale of pathological features we integrate multiple PMP modules for different patch sizes. For evaluation, we have constructed a new dataset, named OPTOS dataset, consisting of 1,033 high-resolution fundus images photographed by Optos camera and conducted comprehensive experiments to validate the efficacy of our proposed method. And the results on both the public dataset and our dataset demonstrate that our method achieves remarkable performance compared to state-of-the-art methods.  
\end{abstract}


\section{Introduction}
\label{sec:introduction}

Retinal diseases, such as diabetic retinopathy (DR), age-related macular degeneration (AMD), and retinopathy of prematurity (ROP), are some of the leading causes of blindness worldwide~\cite{blindness}. Regular retinal examinations can facilitate early disease diagnosis before the manifestation of any symptomatic indications. Timely identification is of paramount significance as it can prevent complete vision loss in patients and potentially impede or halt degenerative conditions through prompt treatment regimens~\cite{selvachandran2023developments}.

 Diagnosing retinal disease often requires comprehensive consideration by clinically experienced ophthalmologists and specialists, a process that is not only very time-consuming and labor-intensive. In populous countries like India, the scarcity of well-trained ophthalmologists poses a significant challenge in addressing the pressing need for comprehensive eye care~\cite{chawla2019retinopathy}. Consequently, there has been a growing reliance on automated analysis and diagnostic systems, such as artificial intelligence-based AI medical screening systems \cite{guo2015computer,saleh2022computer, bourouis2014intelligent}, which not only help alleviate the burden on healthcare personnel but also provide comparable diagnostic outcomes \cite{lee2001comparison}. Particularly, deep learning-based methods for retinal disease diagnosis have been actively investigated.

 Early deep learning-based approaches for diagnosis typically use Convolutional Neural Networks (CNNs) as CNNs have a strong ability to extract features of images. For instance, Asif et al.~\cite{asif2022deep} propose a deep residual network based on the popular CNN architecture ResNet50 for the classification of multiple retinal diseases including DME, choroidal neovascularization (CNV), and DRUSEN. Y.T. et al.~\cite{oh2022end} propose an end-to-end two-branch network based on the EfficientNet, which can automatically classify various retinal diseases and solve problems with severe class imbalance. Pan et al.~\cite{pan2023fundus} design an automated deep learning-based system using Inceptionv3 and ResNet50 for categorizing fundus images into three classes, namely normal, macular degeneration, and tessellated fundus for timely recognition and treatment.

However, CNNs focus on local features and cannot establish long-range connectivity of images. This contradicts the need to focus on the relationship between local lesions and the overall retina in the classification of retinal diseases. Therefore, very recently, another popular deep learning architecture called Vision Transformer (ViT) \cite{dosovitskiy2020image} has been applied for retinal disease diagnosis because its self-attention mechanism can capture long-range associations effectively. For instance, Yu et al.~\cite{yu2021mil} explore the applicability of ViT for retinal disease classification tasks and integrate Multiple Instance Learning into ViT to fully exploit the feature representations in fundus images. Junyong Shen et al.~\cite{shen2023structure} propose a Structure-Oriented ViT (SoT) for retinal disease grading, which can further construct the relationship between lesions and the whole retina. 

Although the based approach focuses on capturing global features, overcoming the weakness of CNNs, it ignores local features to some extent. For some diseases, the lesion area is fixed in the retina, for example, macular degeneration usually occurs in the central region of the retina \cite{Book}. Ignoring such local information may lead to a decrease in disease detection performance. Solving this problem requires the model to balance the ability to capture long-range and local relationships.

Another issue is that, in contrast to the classification of natural objects, accurate multi-class classification of retinal diseases is challenging due to the presence of mild forms and overlapping signs. Conditions such as hard exudate, sub-retinal hemorrhage, neovascularization, pigment epithelial detachment, and macular atrophy can be observed in various retinal diseases including wet age-related macular degeneration (AMD), polypoidal choroidal vasculopathy (PCV), choroidal neovascularization, macular atrophy, retinal angiomatous proliferation, and idiopathic macular telangiectasia~\cite{cen2021automatic}. Thus, it is essential to fully explore the subtle differences between diseases, which requires the model to extract more complex feature representations from retinal images.

In addition, the various sizes of pathological features in fundus images across different diseases is a crucial aspect that has been overlooked in existing related works since certain diseases may only affect a small part of the retina, while others may spread across the entire retina. Thus, the approach aimed at recognizing different diseases should possess the ability to identify and distinguish various scales of lesions.

In this work,  we propose a novel framework named Multi-Scale Patch Message Passing Swin Transformer (PMP-Swin) to address the mentioned challenges. Specifically, we adopt the pre-trained Swin Transformer as the backbone because its special pyramid architecture and shift window-based self-attention enable it to capture both global and local features effectively. Inspired by the fact that lesions of retinal diseases are usually scattered throughout various locations in the fundus image or concentrated in a large area, we design a Patch Message Passing (PMP) module to achieve fine-grained semantic understanding by constructing semantic interactions between each patch of the feature map. Finally, considering the various scales of disease regions, we integrate multiple PMP modules to construct interactions for different patch sizes. We conduct comprehensive experiments on both our private and public datasets to verify the proposed method's effectiveness.
Our key contributions can be summarized as follows:

1) We design a novel PMP module based on the message-passing mechanism to construct semantic associations between patches output by Swin Transformer. The PMP module can effectively help discriminate confusing lesion features, achieving a more accurate diagnosis.

2) To recognize lesions with various scales better, multi-scale PMP modules for different-sized patches are attached to the Swin Transformer.

3) We build a new OPTOS dataset, with 1033 high-resolution colorful fundus images, and show through experiments that our proposed framework PMP-Swin can achieve higher accuracy than previous methods.

\section{RELATED WORK}
\subsection{CNN based Methods}

Extensive methods with CNN architecture have been proposed for retinal disease classification because of the automatic feature extraction ability of CNN networks, which eliminates the need for manual intervention \cite{9524691, rajagopalan2021deep, sunija2021octnet}. Most of the existing works are based on optimizing the classic CNN architectures, such as ResNet proposed by He et al. \cite{he2016deep}, XceptionNet proposed by Chollet \cite{Chollet_2017_CVPR}, for the specific problem of retinal disease diagnosis, thus achieving more accurate diagnosis. For example, Sengar et al. \cite{sengar2023eyedeep} propose a novel deep learning-based multi-layer neural network architecture EyeDeep-Net for the classification of fundus images and non-invasive diagnosis of various eye diseases. Unfortunately, methods based on CNN architectures cannot fully solve the limitation of CNN itself, which is to focus on local attention, and thus cannot effectively extract global semantic information, making precise diagnosis remain challenging. Yang et al.~\cite{yang2022multi} propose a feature extraction network DSRA-CNN based on the Xception architecture to achieve the classification of eight different fundus diseases on the ODIR dataset.

Therefore, some recent works have started to incorporate Self-Attention (SA) \cite{8882308, 10222689, wang2023combining} into CNN architectures for diagnosis to add global attention rather than just local attention. Particularly, Wang et al. \cite{wang2023combining} propose a multi-level fundus image classification model MBSaNet that combines CNN and Self-Attention mechanisms. The convolutional block extracts local information of the fundus image, and the SA module further captures the complex relationship between different spatial positions, thereby directly detecting one or more fundus diseases in the fundus image. Experimental results show that MBSaNet achieves state-of-the-art performance with fewer parameters on two different datasets. These works demonstrate that the attention mechanism is promising in disease classification diagnosis.

\subsection{Transformer based Methods}

Due to the excellent performance of the vanilla Transformer in various computer vision tasks, many recent works \cite{jiang2022satformer, ma2022hctnet, playout2022focused, yu2021mil, yang2021fundus} have explored the effectiveness of the Transformer architecture to address the limitation of CNNs in fundus image classification. Especially, N. S. Kumar et al.~\cite{kumar2021diabetic} evaluate the DR grading architectures of Transformer, CNN, and Multi-Layer Perceptron (MLP) in terms of model convergence time, accuracy, and model scale, demonstrating that the Transformer-based model outperforms the CNN and MLP architectures in terms of not only accuracy but also achieving comparable model convergence time. 

Currently, methods based on the Transformer backbone for retinal disease diagnosis can be roughly divided into two types: using only the Transformer and combining the Transformer with CNN for extracting features. For the former, notably, Yu et al. \cite{yu2021mil} first use ViT for retinal disease classification tasks by pre-training the Transformer model for downstream retinal disease classification tasks. In addition, to fully utilize the feature representation extracted from a single image patch, they propose a MIL head based on multi-instance learning (MIL). They test on DR grading dataset APTOS2019 and dataset RFMiD2020, respectively, and achieve state-of-the-art performance. Similarly, Jiang et al. \cite{jiang2022satformer} also verify the effectiveness of the ViT backbone for classifying retinal disease. They design an innovative saliency enhancement module and abnormality-aware attention to distinguish the main abnormal regions as well as the small subtle lesions for better diagnosis. For the latter, Yang et al. \cite{yang2021fundus} propose a method called TransEye which combines the advantages of CNN and ViT, enabling effective extraction of bottom-level features and establishing long-range dependencies in the images. Experimental evaluations conducted on the OIA-ODIR dataset demonstrated that the TransEye method achieves much higher optimal prediction accuracy compared to CNN and ViT. From the identified Transformer-related works, it can be observed that the Transformer has great potential to achieve excellent performance in retinal disease diagnosis tasks, surpassing most CNNs. The current challenge is how to innovate upon the Transformer-based architecture in order to make it more applicable for retinal disease diagnosis.

\begin{figure*}[t]
\centerline{\includegraphics[width=\linewidth]{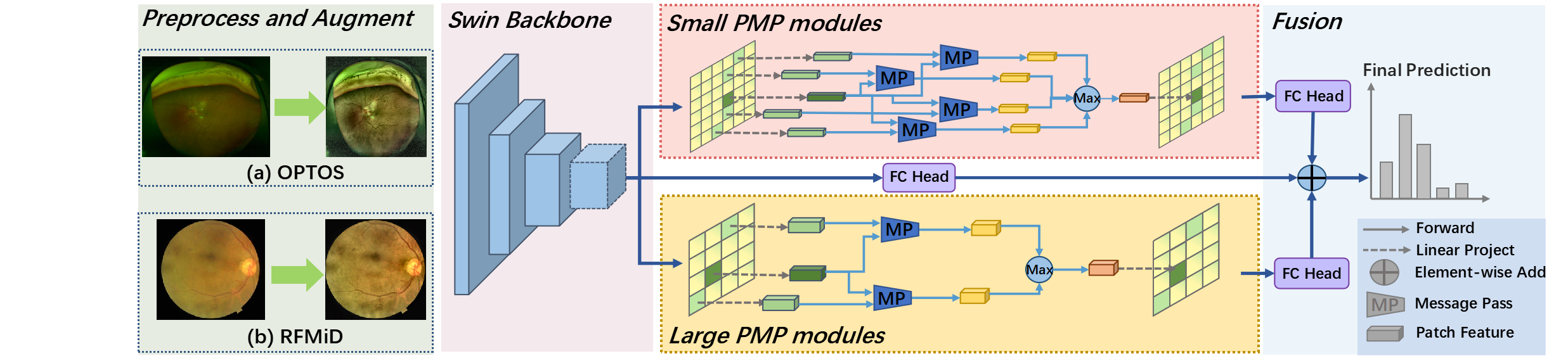}}
\caption{The framework of our proposed method. Firstly, the retinal images undergo preprocessing and data augmentation and are then fed into the Swin Transformer backbone to obtain a semantic feature map. This map is then linearly transformed and inputted into two branches, namely PMP modules for small patches and PMP modules for large patches, to obtain new feature maps. The three resulting semantic feature maps are fused and used to compute the final prediction. }
\label{framework}
\end{figure*}

\section{METHOD}

\subsection{Overview}

The proposed method is illustrated in Fig. \ref{framework}. After preprocessing and data augmentation, the input image is fed into the Swin V2 \cite{liu2022swin} backbone to obtain a feature map consisting of many patches.  The semantic feature map is then directed into dual branches of PMP modules to establish global connections between semantic features of different lesion scales through patch message aggregation. The output results of the two branches and the main trunk classifier will be used separately to calculate the Cross-Entropy loss.


\subsection{Preprocessing and Data Augmentation}

Our proposed method has been evaluated using two distinct datasets. 
\subsubsection{OPTOS Dataset}
The first dataset, OPTOS, comprises 1033 high-definition color fundus images captured using advanced ultra-widefield cameras manufactured by Optos. These images cover more than 80\% or $200^{\circ}$ of the retina in a single capture. The dataset is categorized by several experienced ophthalmologists into seven classes: DME, High Myopia, Hypertension, Uveitis, Retinal Vein Occlusion (RVO), Retinal Detachment, and Normal, which is shown in Table \ref{OPTOS}. The images have varying resolutions ranging from 714 $\times$ 545 to 3900 $\times$ 3072. To facilitate comparison with other methods, all images were resized to a standardized scale of 384 $\times$ 384. We augment the datasets using various data augmentation techniques to mitigate the adverse effects of class imbalance on model performances. Specifically, we apply CLAHE, Gaussian blur, horizontal and vertical flips, affine transformations, and color jittering to achieve a balanced distribution. 

\begin{figure}[H]
\centerline{\includegraphics[width=\columnwidth]{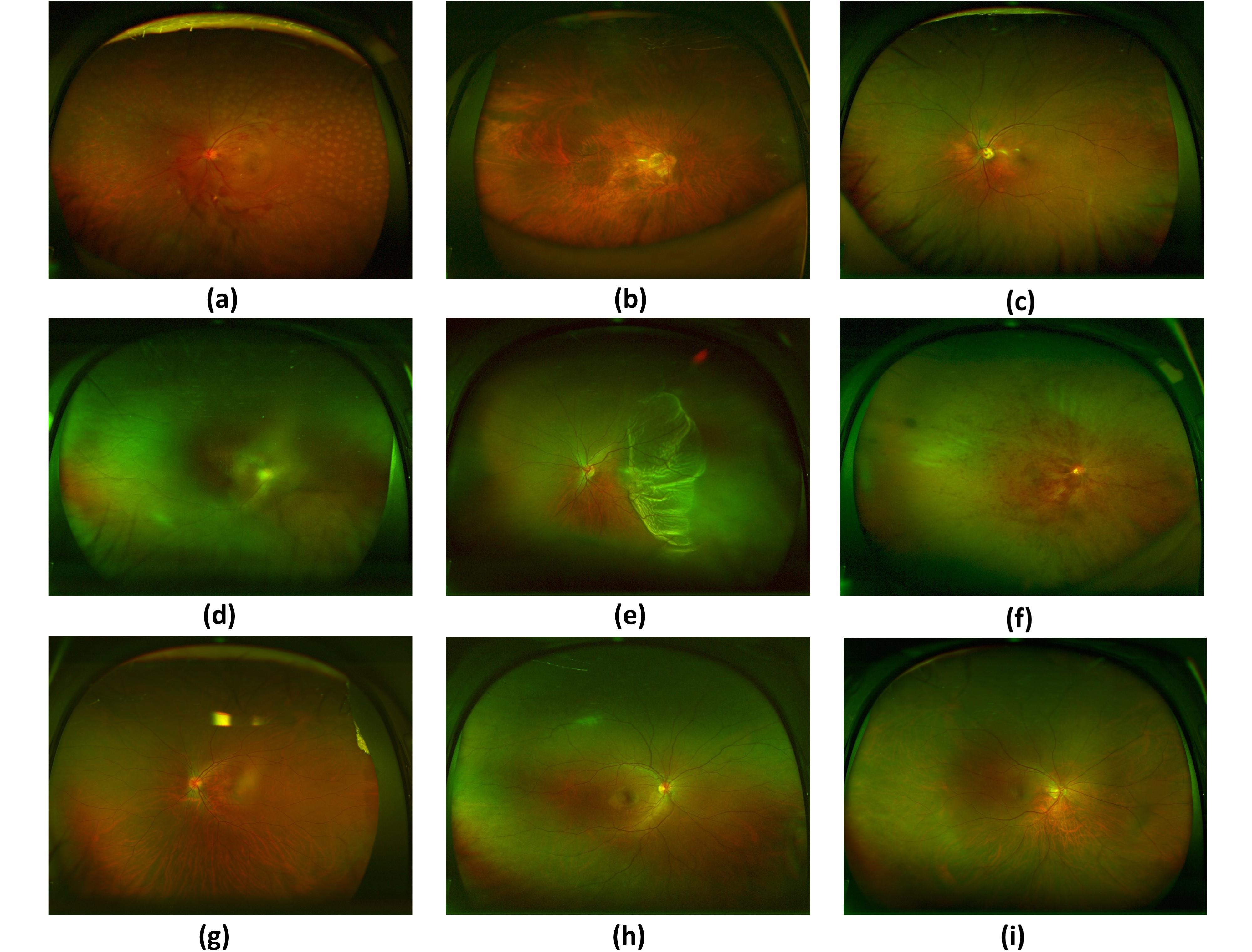}}
\caption{Sample images in OPTOS dataset, \textbf{a}: DME, \textbf{b}: High Myopia, \textbf{c}: Hypertension, \textbf{d}: Uveitis, \textbf{e}: Retinal Detachment, \textbf{f}: RVO, \textbf{g-i}: Normal.}
\label{samples}
\end{figure}

\begin{table}[ht]
\caption{Distribution of OPTOS Dataset}
\centering
\begin{tabular}{c|ccc|ccc}
\hline
\multirow{2}{*}{Catagory} & \multicolumn{3}{c|}{Unbalanced}                              & \multicolumn{3}{c}{Balanced}                                 \\ \cline{2-7} 
                          & \multicolumn{1}{c|}{Train} & \multicolumn{1}{c|}{Val} & Test & \multicolumn{1}{c|}{Train} & \multicolumn{1}{c|}{Val} & Test \\ \hline
DME                       & \multicolumn{1}{c|}{49}    & \multicolumn{1}{c|}{12}  & 16   & \multicolumn{1}{c|}{160}   & \multicolumn{1}{c|}{40}  & 50   \\ \hline
High Myopia               & \multicolumn{1}{c|}{156}   & \multicolumn{1}{c|}{40}  & 49   & \multicolumn{1}{c|}{160}   & \multicolumn{1}{c|}{40}  & 50   \\ \hline
Hpyertension              & \multicolumn{1}{c|}{64}    & \multicolumn{1}{c|}{16}  & 20   & \multicolumn{1}{c|}{160}   & \multicolumn{1}{c|}{40}  & 50   \\ \hline
RVO                       & \multicolumn{1}{c|}{133}   & \multicolumn{1}{c|}{34}  & 42   & \multicolumn{1}{c|}{160}   & \multicolumn{1}{c|}{40}  & 50   \\ \hline
Retinal Detachment        & \multicolumn{1}{c|}{20}    & \multicolumn{1}{c|}{16}  & 63   & \multicolumn{1}{c|}{160}   & \multicolumn{1}{c|}{40}  & 50   \\ \hline
Uveitis                   & \multicolumn{1}{c|}{130}   & \multicolumn{1}{c|}{33}  & 41   & \multicolumn{1}{c|}{160}   & \multicolumn{1}{c|}{40}  & 50   \\ \hline
Normal                    & \multicolumn{1}{c|}{63}    & \multicolumn{1}{c|}{16}  & 20   & \multicolumn{1}{c|}{160}   & \multicolumn{1}{c|}{40}  & 50   \\ \hline
\multirow{2}{*}{\textbf{Total}} & \multicolumn{1}{c|}{615}   & \multicolumn{1}{c|}{167} & 251  & \multicolumn{1}{c|}{1120}  & \multicolumn{1}{c|}{280} & 350  \\ \cline{2-7} 
                                & \multicolumn{3}{c|}{1033}                                     & \multicolumn{3}{c}{1750}                                     \\ \hline

\end{tabular}
\label{OPTOS}
\end{table}

\subsubsection{RFMiD Dataset}
Additionally, we utilize an open-source multi-labeled Retinal Fundus Multi-Disease Image Dataset (RFMiD) ~\cite{pachade2021retinal} for the second dataset. This dataset, collected from the Eye Clinic of Sushrusha Hospital in collaboration with the Centre of Excellence in Signal and Image Processing, SGGS Institute of Engineering and Technology, India, comprises 1920 color fundus images with 46 different categories. As RFMiD is a multi-labeled dataset, each fundus image may belong to more than one disease category. Due to the limited number of images in certain categories, which is insufficient for training deep learning models, single-label images with a count of at least 100 are selected from each category to ensure a balanced distribution of data following the previous method \cite{sengar2023eyedeep}. This results in the selection of four categories, namely DR, MH, ODC, and Normal, shown in Table \ref{RFMiD}. Subsequently, we apply the same preprocessing and data augmentation techniques as employ our own dataset to balance the distribution in RFMiD. Fig. \ref{framework} displays the results of the preprocessing and data augmentation techniques.
\begin{table}[ht]
\caption{Distribution of RFMiD Dataset}
\centering
\begin{tabular}{c|ccc|ccc}
\hline
\multirow{2}{*}{Catagory}       & \multicolumn{3}{c|}{Unbalanced}                              & \multicolumn{3}{c}{Balanced}                                 \\ \cline{2-7} 
                                & \multicolumn{1}{c|}{Train} & \multicolumn{1}{c|}{Val} & Test & \multicolumn{1}{c|}{Train} & \multicolumn{1}{c|}{Val} & Test \\ \hline
DME                             & \multicolumn{1}{c|}{163}   & \multicolumn{1}{c|}{41}  & 36   & \multicolumn{1}{c|}{306}   & \multicolumn{1}{c|}{76}  & 68   \\ \hline
MH                              & \multicolumn{1}{c|}{122}   & \multicolumn{1}{c|}{31}  & 27   & \multicolumn{1}{c|}{306}   & \multicolumn{1}{c|}{76}  & 68   \\ \hline
ODC                             & \multicolumn{1}{c|}{80}    & \multicolumn{1}{c|}{21}  & 18   & \multicolumn{1}{c|}{306}   & \multicolumn{1}{c|}{76}  & 68   \\ \hline
Normal                          & \multicolumn{1}{c|}{272}   & \multicolumn{1}{c|}{68}  & 61   & \multicolumn{1}{c|}{306}   & \multicolumn{1}{c|}{76}  & 68   \\ \hline
\multirow{2}{*}{\textbf{Total}} & \multicolumn{1}{c|}{637}   & \multicolumn{1}{c|}{168} & 142  & \multicolumn{1}{c|}{1224}  & \multicolumn{1}{c|}{304} & 272  \\ \cline{2-7} 
                                & \multicolumn{3}{c|}{947}                                     & \multicolumn{3}{c}{1800}                                     \\ \hline
\end{tabular}
\label{RFMiD}
\end{table}

\subsection{Message Passing Module}

\begin{algorithm}[ht]
\label{MPMP}
\caption{Multi-scale Patch Message Passing}
\LinesNumbered
\KwIn{
$\mathcal{M}$: Semantic feature map output by Swin backbone \newline
$S_{s}$: Size of small patch \newline
$S_{k}$: $k$ nearest neighbor patches of small branch \newline
$S_{n}$: Number of small PMP modules \newline
$L_{s}$: Size of large patch \newline
$L_{k}$: $k$ nearest neighbor patches of large branch \newline
$L_{n}$: Number of large PMP modules
}
\KwOut{
$\mathcal{F}$: Final prediction
}
Regarding the primary branch, $\mathcal{M}$ is transmitted through the fully connected layer, producing the logit value of $y^{'}$:
\centerline{$y^{'} = FC(\mathcal{M})$;} 

For the branch of the large patch
\centerline{$P_{L} = LinearProject(\mathcal{M}, L_{s}) = \left \{ p_{1}, p_{2}, \dots, p_{L_{m}} \right \}$;} 
\For{$1$ to $L_{n}$}{
\For{$p_{i}$ $\gets $ $p_{1}$ to $P_{L_{m}}$}{
    Propagate messages from the $L_{k}$ nearest neighbors in the feature space
    $\left \{p_{j1}, p_{j2}, \dots, p_{jL_{k}} \right \} = KNearest(P_{L}, p_{i})$ \;
    $p_{i} = \mathop{max}\limits_{ p_{j} \in \left \{p_{j1}, p_{j2}, \dots, p_{jL_{k}} \right \}}  Message(p_{i}, p_{j})$ \;
}
}
Calculate the logit $y^{''}$ output by the branch of the large patch
\centerline{$y^{''} = FC(P_{L})$;}

For the branch of the small patch
\centerline{$P_{S} = LinearProject(\mathcal{M}, S_{s}) = \left \{ p_{1}, p_{2}, \dots, p_{S_{m}} \right \}$;}
\For{$1$ to $S_{n}$}{
\For{$p_{i}$ $\gets $ $p_{1}$ to $P_{S_{m}}$}{
    Propagate messages from the $S_{k}$ nearest neighbors in the feature space
    $\left \{p_{j1}, p_{j2}, \dots, p_{jS_{k}} \right \} = KNearest(P_{S}, p_{i})$ \;
    $p_{i} = \mathop{max}\limits_{ p_{j} \in \left \{p_{j1}, p_{j2}, \dots, p_{jS_{k}} \right \}}  Message(p_{i}, p_{j})$ \;
}
}
Calculate the logit $y^{'''}$ output by the branch of the small patch
\centerline{$y^{'''} = FC(P_{S})$;}

$\mathcal{F} = Softmax(Mean(y^{'}, y^{''}, y^{'''}))$ \;
\Return $\mathcal{F}$
\end{algorithm}

When it comes to classification tasks, it is common practice to feed the entire feature map into a linear classifier to get the final logits. However, this method can result in the loss of semantic information on lesion regions in fundus images. To make the most of the feature representations of the feature map, we utilize patches and input them into our specially designed PMP modules to create semantic associations.

Initially, the input image $X \in R^{H \times W \times 3 }$ is divided into a collection of non-overlapping patches with size $4 \times 4$  by patch partition, where $H, W, 3$ denotes the height, width and the number of channels, respectively. Then, these patches are fed into the Transformer blocks, generating the feature map with a dimension of $\frac{H}{32} \times \frac{W}{32} \times 8C$ ~\cite{liu2021swin}, consisting of $\frac{H}{32} \times \frac{W}{32}$ patches with a dimension of $1 \times 1 \times 8C$.

\begin{figure}[ht]
\centerline{\includegraphics[width=\linewidth]{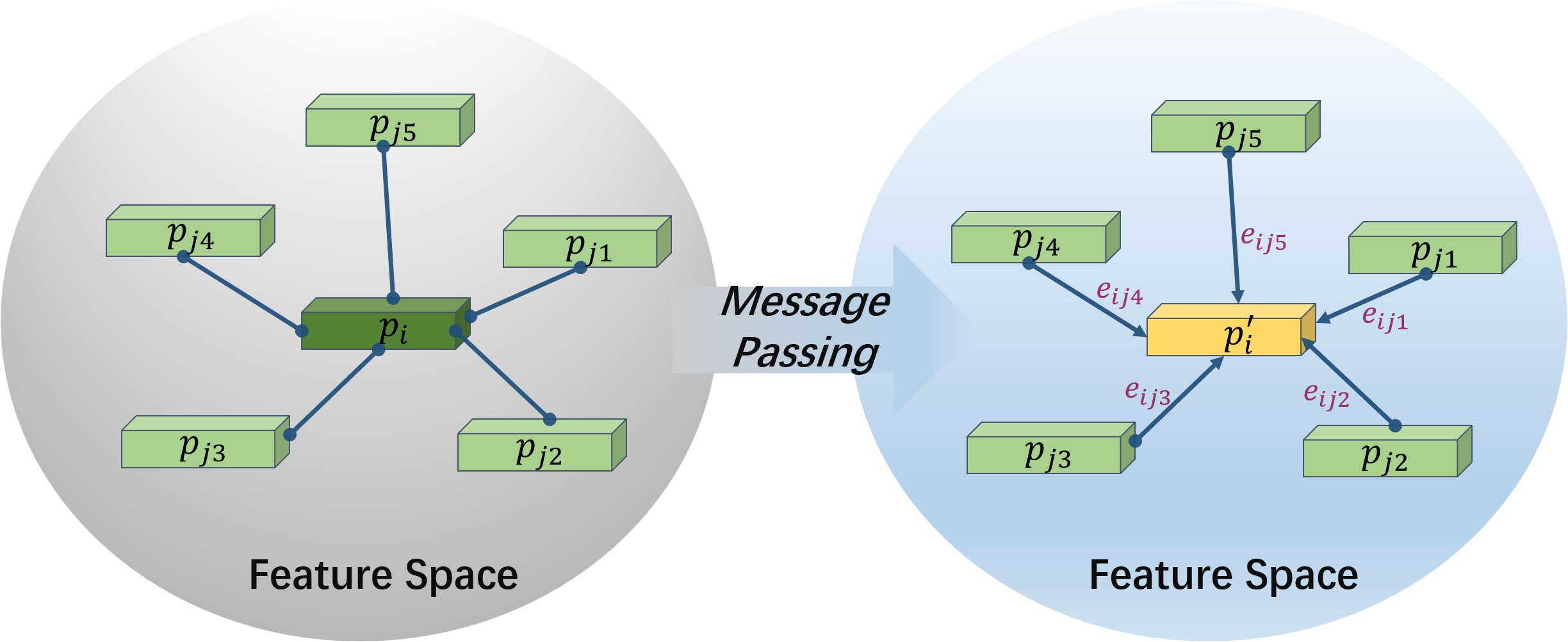}}
\caption{Patch message passing. Each patch's feature is represented by $p$, and the connections between patches indicate their adjacency in feature space. The message transmitted between two patches is represented by $e$.}
\label{pass}
\end{figure}

Specifically, inspired by Message Passing Neural Networks (MPNNs)~\cite{gilmer2017neural, wang2019dynamic}, we adapt graph convolution operations to transmit and aggregate information between patches with similar features, which allows patches with receptive fields corresponding to the lesion regions to directly establish semantic relationships dynamically. Showed in Fig. \ref{pass}, we consider each patch feature as a graph node $p$, and then the feature map with $n$ patches can be represented as $P = \left \{ p_{1},\dots,p_{n}\right \}  \subseteq R^{1 \times 1 \times F}$. $F$ denotes the feature dimension of each patch. During the subsequent message passing, patches with similar semantic features are connected to build a graph using k-nearest neighbors in the feature space, i.e., we construct the directed graph $G = (V, E)$, where vertices $V = \left \{ 1, 2, \dots, n \right \}$ and edges $E$ are defined as:
\begin{equation}
E = \left \{ e_{ij} \mid i,j\in \left \{ 1, 2, \dots, n\right \}, i\ne j \right \} \subseteq V \times V.
\end{equation} Here, $e_{ij}$ represents the edge connecting $p_{i}$ and $p_{j}$ as a nonlinear mapping of the semantic information of the two patches, which is defined as:
\begin{equation}
    e_{ij} = h_{\Theta }\left ( p_{i}, p_{j} - p_{i}\right ),
\end{equation} where $h_{\Theta}$ denotes a neural network with a set of learnable parameters $\Theta$. It can be defined as:
\begin{equation}
\begin{split}
& h_{\Theta}\left ( \cdot \right ) = \\
& Dropout\left ( LeakyReLU\left ( LayerNorm\left ( Linear \left ( \cdot \right) \right )  \right )  \right ),
\end{split}
\end{equation}
where $Linear$ represents a linear layer responsible for transforming the input data into a new representation, $LayerNorm$ represents a layer-normalization layer that helps to improve the stability and efficiency of the network, $LeakyReLU$ is an activation function that introduces nonlinearity into the model, and $Dropout$ represents a dropout layer that serves to mitigate the risk of overfitting by randomly dropping out specific units during training. After the message aggregation, each patch is updated to:
\begin{equation}
p'_{i} = \mathop{max}\limits_{j:e_{ij}\in E } e_{ij}, i \in \left \{ 1, 2, \dots, n\right \}. 
\end{equation}
 We stack multiple PMP modules to enable dynamic interaction of pathological semantic features, thus effectively establishing global relationships.

\subsection{Multi-Scale Patch}
Due to the varied location and size of lesion areas in different retinal diseases, if the patch size is too small, it is likely to aggregate information from similar patches rather than neighboring patches, which can lead to a convolution-like operation and hinder the establishment of global connections. To fuse information from patches over long distances, it is necessary to perform multiple information aggregation operations on individual patches, which results in higher FLOPs and memory consumption. Conversely, if the patch size is too large, it is difficult to extract detailed image information. Therefore, our proposed approach aims to capitalize on the benefits of utilizing smaller patch sizes, while also ensuring that the overall complexity is properly balanced. Specifically, we have added two PMP module branches to the Transformer, targeting different patch sizes, as described in Algorithm 1. As for the large branch, we concatenate adjacent $L_{s}$ patches in the feature dimension to represent a larger patch. Therefore, a total of $ \frac { H } { 32 \times L_{s} } \times \frac { W } { 32 \times L_{s} } $ patches are inputted into the PMP modules of the large branch. Similarly, after the linear transform, a total of $ \frac { H } { 32 \times S_{s} } \times \frac { W } { 32 \times S_{s} } $ patches are sent into corresponding PMP modules. Finally, we obtain three feature maps including $\mathcal{M}$, $P_{L}$, and $P_{S}$, which we feed into the FC head to obtain corresponding logits. We calculate the loss for each of the three logits separately using CE loss. Our loss function can be denoted as:

\begin{equation}
    L\left ( y,y' ,y'',y''' \right ) = -\frac{1}{3} \sum_{i=0}^{C} y_{i}log\left ( y'_{i} y''_{i} y'''_{i} \right ),
\end{equation}
where $y$ represents the ground truth and $C$ represents the number of categories. The primary branch, large branch, and small branch output logits are represented by $y'$, $y''$, and $y'''$, respectively.

\section{EXPERIMENTAL SETUP AND RESULTS}

\subsection{Metrics}

The Performance evaluation is based on four metrics, namely Accuracy, Precision, F1, and Cohen's Kappa, which are given by:


\begin{equation}
Accuracy = \frac{True \: Positive + True \: Negative}{Positive + Negative}
\end{equation}

\begin{equation}
Precision = \frac{True \: Positive }{ True \: Positive + False \: Postive}
\end{equation}

\begin{equation}
Recall = \frac{True \: Positive }{ True \: Positive + False \: Negative}   
\end{equation}

\begin{equation}
F1 = 2\ast \frac{Precision \ast Recall}{Precision+Recall} 
\end{equation}

\begin{equation}
Kappa = \frac{P_{o} -P_{e}}{1 - P_{e}}
\end{equation} 

\begin{equation}
P_{o} = \frac{True \: Positive + True \: Negative}{Positive + Negative}  = Accuracy 
\end{equation}

\begin{equation}
P_{e} = \frac{  \sum_{i=1}^{C} \left (  a_{i} \times b_{i}\right ) }{ N \times N    },
\end{equation}

where $a_{i}$ refers to the number of the practical samples for class $i$, $b_{i}$ refers to the number of the predicted samples for class $i$, $C$ is the number of classes and $N$ is the number of total samples.

\begin{table*}[t]
\caption{Comparison of Results Obtained By Different Methods On OPTOS Dataset}
\centering
\resizebox{\linewidth}{!}{
\begin{tabular}{c|c|cccc|cccc}
\hline
\multirow{2}{*}{Category} &
  \multirow{2}{*}{Method} &
  \multicolumn{4}{c|}{Unbalanced} &
  \multicolumn{4}{c}{Balanced} \\ \cline{3-10} 
 &
   &
  \multicolumn{1}{c|}{Accuracy(\%)} &
  \multicolumn{1}{c|}{Precision(\%)} &
  \multicolumn{1}{c|}{F1(\%)} &
  Kappa(\%) &
  \multicolumn{1}{c|}{Accuracy(\%)} &
  \multicolumn{1}{c|}{Precision(\%)} &
  \multicolumn{1}{c|}{F1(\%)} &
  Kappa(\%) \\ \hline
\multirow{4}{*}{CNN} &
ResNet18 ~\cite{he2016deep} &
  \multicolumn{1}{c|}{72.44 $\pm$ 2.16} &
  \multicolumn{1}{c|}{72.32 $\pm$ 0.31} &
  \multicolumn{1}{c|}{70.74 $\pm$ 1.01 } &
  67.19 $\pm$ 2.22 &
  \multicolumn{1}{c|}{87.60 $\pm$ 1.66} &
  \multicolumn{1}{c|}{88.02 $\pm$ 1.71} &
  \multicolumn{1}{c|}{87.59 $\pm$ 1.72} &
  85.53 $\pm$ 2.26 \\ \cline{2-10} 
 &
  ResNet50~\cite{he2016deep} &
  \multicolumn{1}{c|}{71.95 $\pm$ 1.01} &
  \multicolumn{1}{c|}{71.36 $\pm$ 1.29} &
  \multicolumn{1}{c|}{70.61 $\pm$ 1.68} &
  66.61 $\pm$ 1.47 &
  \multicolumn{1}{c|}{89.49 $\pm$ 1.53} &
  \multicolumn{1}{c|}{89.64 $\pm$ 1.36} &
  \multicolumn{1}{c|}{89.48 $\pm$ 1.56} &
  87.73 $\pm$ 2.08\\ \cline{2-10} 
 &
  SE-ResNet~\cite{hu2018squeeze}&
  \multicolumn{1}{c|}{76.25 $\pm$ 3.99} &
  \multicolumn{1}{c|}{73.61 $\pm$ 9.13} &
  \multicolumn{1}{c|}{73.62 $\pm$ 6.72} &
  71.61 $\pm$ 5.12 &
  \multicolumn{1}{c|}{90.00 $\pm$ 0.94} &
  \multicolumn{1}{c|}{90.42 $\pm$ 0.68} &
  \multicolumn{1}{c|}{89.90 $\pm$ 0.84 } &
  88.33 $\pm$ 1.27 \\ \cline{2-10} 
 &
  MobileNet \cite{DBLP:journals/corr/abs-1905-02244} &
  \multicolumn{1}{c|}{74.90 $\pm$ 2.70} &
  \multicolumn{1}{c|}{73.25 $\pm$ 1.52} &
  \multicolumn{1}{c|}{72.29 $\pm$ 2.66} &
  70.00 $\pm$ 3.44 &
  \multicolumn{1}{c|}{88.97 $\pm$ 3.37} &
  \multicolumn{1}{c|}{89.50 $\pm$ 3.46} &
  \multicolumn{1}{c|}{89.07 $\pm$ 3.59} &
  87.13 $\pm$ 4.59\\ \hline
\multirow{5}{*}{Transformer} &
ViT~\cite{dosovitskiy2020image} &
  \multicolumn{1}{c|}{74.81 $\pm$ 3.53} &
  \multicolumn{1}{c|}{74.36 $\pm$ 3.84} &
  \multicolumn{1}{c|}{72.32 $\pm$ 3.18} &
  69.81 $\pm$ 4.30 &
  \multicolumn{1}{c|}{89.26 $\pm$ 1.08} &
  \multicolumn{1}{c|}{89.64 $\pm$ 1.01} &
  \multicolumn{1}{c|}{89.33 $\pm$ 1.13} &
  87.47 $\pm$ 1.48\\ \cline{2-10} 
 &
  Cross-ViT ~\cite{chen2021crossvit} &
  \multicolumn{1}{c|}{74.57 $\pm$ 7.42} &
  \multicolumn{1}{c|}{74.83 $\pm$ 7.16} &
  \multicolumn{1}{c|}{72.36 $\pm$ 10.67} &
  68.95 $\pm$ 6.82&
  \multicolumn{1}{c|}{88.46 $\pm$ 2.19} &
  \multicolumn{1}{c|}{88.26 $\pm$ 4.23} &
  \multicolumn{1}{c|}{88.48 $\pm$ 2.12} &
  86.96 $\pm$ 2.96 \\ \cline{2-10} 
 &
  MIL-VT~\cite{yu2021mil}&
  \multicolumn{1}{c|}{75.45 $\pm$ 6.40 } &
  \multicolumn{1}{c|}{74.06 $\pm$ 9.76} &
  \multicolumn{1}{c|}{75.37 $\pm$ 8.76} &
  73.19 $\pm$ 6.97 &
  \multicolumn{1}{c|}{88.40 $\pm$ 1.08} &
  \multicolumn{1}{c|}{88.67 $\pm$ 1.51} &
  \multicolumn{1}{c|}{88.41 $\pm$ 1.38} &
  86.47 $\pm$ 1.48 \\ \cline{2-10} 
 &
  Swin~\cite{liu2022swin} &
  \multicolumn{1}{c|}{77.59 $\pm$ 2.38 } &
  \multicolumn{1}{c|}{75.20 $\pm$ 1.95 } &
  \multicolumn{1}{c|}{75.37 $\pm$ 2.60} &
  73.19 $\pm$ 3.45 &
  \multicolumn{1}{c|}{90.51 $\pm$ 3.19} &
  \multicolumn{1}{c|}{90.76 $\pm$ 2.30 } &
  \multicolumn{1}{c|}{90.53 $\pm$ 2.96 } &
  88.93 $\pm$ 4.36 \\ \cline{2-10} 
 &
  \textbf{PMP-Swin(Ours)} &
  \multicolumn{1}{c|}{\textbf{80.29 $\pm$ 1.04}} &
  \multicolumn{1}{c|}{\textbf{78.60 $\pm$ 0.75}} &
  \multicolumn{1}{c|}{\textbf{78.00 $\pm$ 0.79}} &
  \textbf{76.36 $\pm$ 1.38} &
  \multicolumn{1}{c|}{\textbf{92.12 $\pm$ 0.27}} &
  \multicolumn{1}{c|}{\textbf{92.29 $\pm$ 0.36}} &
  \multicolumn{1}{c|}{\textbf{92.13 $\pm$ 0.29}} &
  \textbf{90.80 $\pm$ 0.37} \\ \hline
\end{tabular}
}
\label{table_1}
\end{table*}

\begin{figure*}[t]
\centerline{\includegraphics[width=\linewidth]{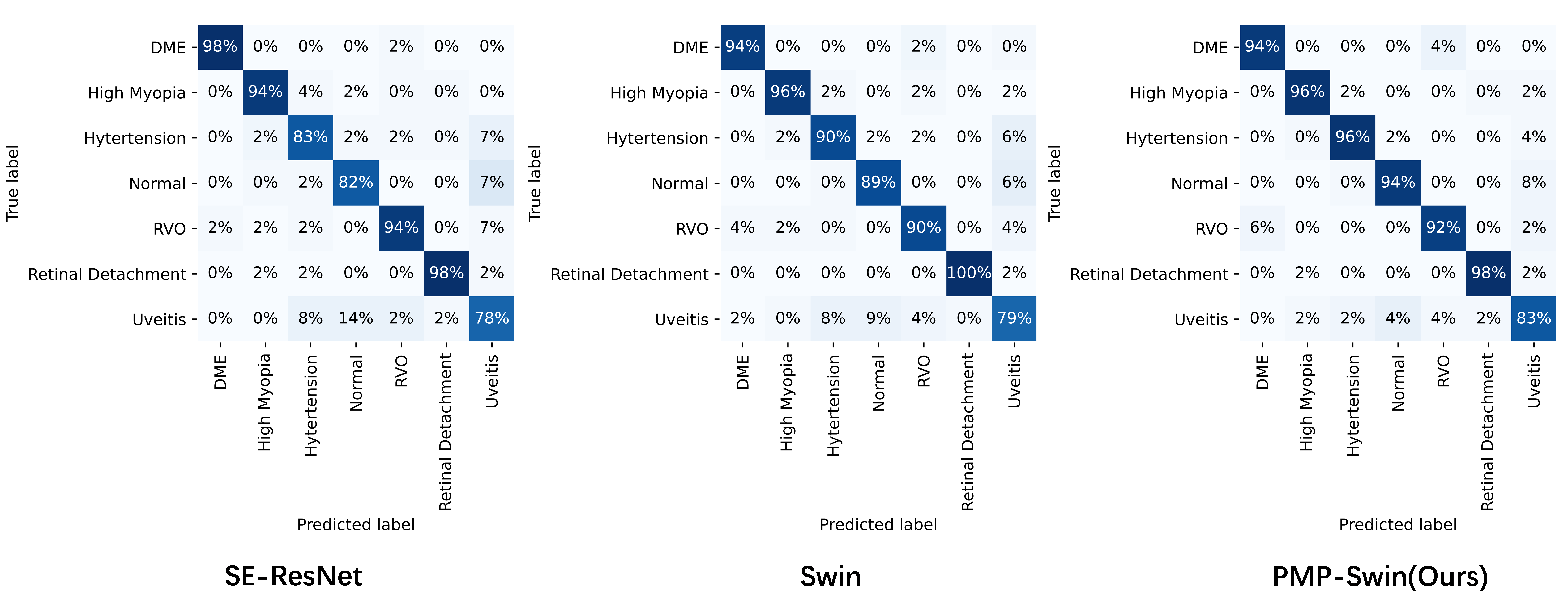}}
\caption{Comparison of confusion matrices obtained from different methods on OPTOS dataset. The column and row denote the predicted and true labels, respectively. The intensity of the matrix entries is proportional to the magnitude of the corresponding values, with darker shades indicating higher values.}
\label{fig9}
\end{figure*}

\begin{table*}[t]
\caption{Comparison of Results Obtained By Different Methods On Balanced RFMiD Dataset}
\centering
\begin{tabular}{c|c|c|c|c|c}
\hline
Category                     & Method    & Accuracy(\%)       & Precision(\%)      & F1(\%)             & Kappa(\%)          \\ \hline
\multirow{4}{*}{CNN}         & ResNet18~\cite{he2016deep}  & 96.40 $\pm$ 0.40        & 96.45 $\pm$ 0.40        & 96.40 $\pm$ 0.42        & 95.20 $\pm$ 0.72        \\ \cline{2-6} 
                             & ResNet50~\cite{he2016deep}  & 95.15 $\pm$ 1.76          & 95.26 $\pm$ 1.62        & 95.13 $\pm$ 1.77        & 93.53 $\pm$ 3.12        \\ \cline{2-6} 
                             & SE-ResNet~\cite{hu2018squeeze}  & 96.40 $\pm$ 0.91        & 96.46 $\pm$ 0.85        & 96.40 $\pm$ 0.90        & 95.20 $\pm$ 1.61        \\ \cline{2-6} 
                             & MobileNet\cite{DBLP:journals/corr/abs-1905-02244}  & 95.37 $\pm$ 0.44         & 95.45 $\pm$ 0.37        & 95.36 $\pm$ 0.44        & 93.83 $\pm$ 0.79        \\ \hline
\multirow{5}{*}{Transformer} & ViT~\cite{dosovitskiy2020image}        & 95.96 $\pm$ 0.95        & 96.01 $\pm$ 0.89        & 95.96 $\pm$ 0.95        & 94.61 $\pm$ 1.68        \\ \cline{2-6} 
                             & Cross-ViT \cite{chen2021crossvit}  & 95.37 $\pm$ 0.51        & 95.03 $\pm$ 2.60        & 95.40 $\pm$ 0.43        & 94.22 $\pm$ 0.05        \\ \cline{2-6} 
                             & MIL-VT~\cite{yu2021mil}    & 97.06 $\pm$ 0.61        & 97.10 $\pm$ 0.60        & 97.07 $\pm$ 0.62         & 96.08 $\pm$ 1.08        \\ \cline{2-6} 
                             & Swin~\cite{liu2022swin}      & 96.47 $\pm$ 1.12        & 96.53 $\pm$ 1.12         & 96.47 $\pm$ 1.12        & 95.29 $\pm$ 2.00        \\ \cline{2-6} 
                             & \textbf{PMP-Swin(Ours)}  & \textbf{97.79 $\pm$ 0.47} & \textbf{97.83 $\pm$ 0.46} & \textbf{97.80 $\pm$ 0.46} & \textbf{97.06 $\pm$ 0.84}\\ \hline
\end{tabular}
\label{table_2}
\end{table*}

\subsection{Implementation Details}
Using Python based on the PyTorch platform \cite{NEURIPS2019}, all methods are implemented on 4 Nvidia GeForce RTX 3090Ti GPUs, each with 24GB memory. The input images are first resized to a standard size of 384 × 384 and then undergo random augmentation using Python's Albumentations library \cite{info11020125}. The optimizer utilized is Adam, with the learning rate updated using cosine annealing. All model backbones are initialized with pre-trained ImageNet \cite{deng2009imagenet} weights to enhance performance and speed up convergence. The division of the training and testing sets is shown in Table \ref{OPTOS} and \ref{RFMiD}, with the proportion of OPTOS and the public dataset being 0.8:0.2 and 0.85:0.15, respectively. To comprehensively evaluate the performance, we employ a five-fold cross-validation method and calculate the average results of the five models on the testing set. For all methods, we standardize the training process by conducting 150 epochs with a batch size of 8.

\subsection{Experimental Results}

In order to compare with previous methods, we conducted relevant work. Most latest methods for fundus disease classification are difficult to re-implement because of the lack of publicly available code, experimental details, or certain datasets \cite{rodriguez2022multi}. Therefore, in addition to the methods specifically designed for fundus image classification, such as MIL-VT \cite{yu2021mil}, we also select models commonly used as fundus disease classification benchmarks, such as ResNet18 and ResNet50 \cite{he2016deep}. We divide them into CNN and Transformer methods based on their backbones.

\begin{figure}[t]
\centerline{\includegraphics[width=\linewidth]{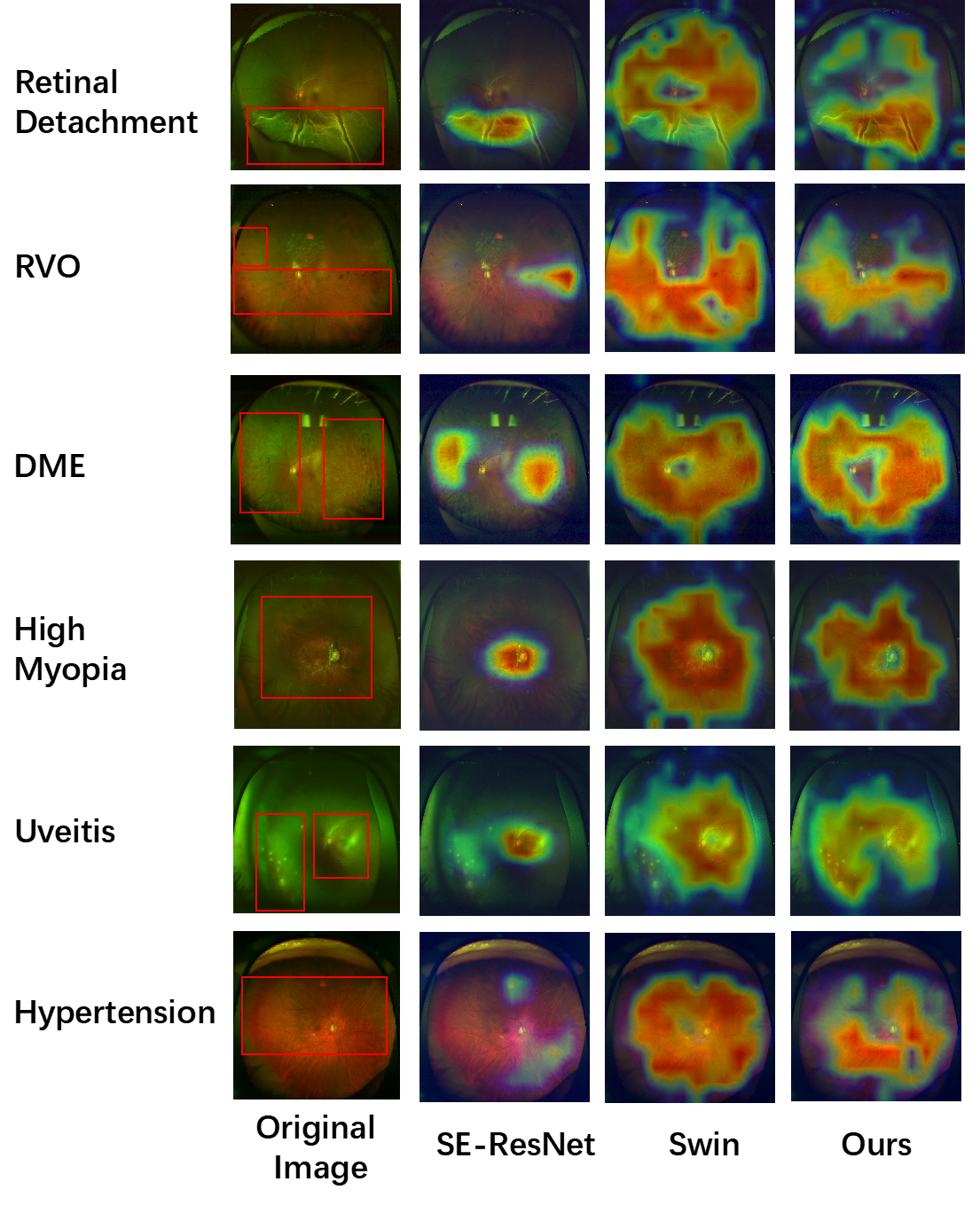}}
\caption{Comparison of Grad-CAM heatmaps of each class obtained by different methods on OPTOS dataset. Lesion location is indicated by the red box.}
\label{heatmap}
\end{figure}

\begin{figure}[t]
\centerline{\includegraphics[width=\linewidth]{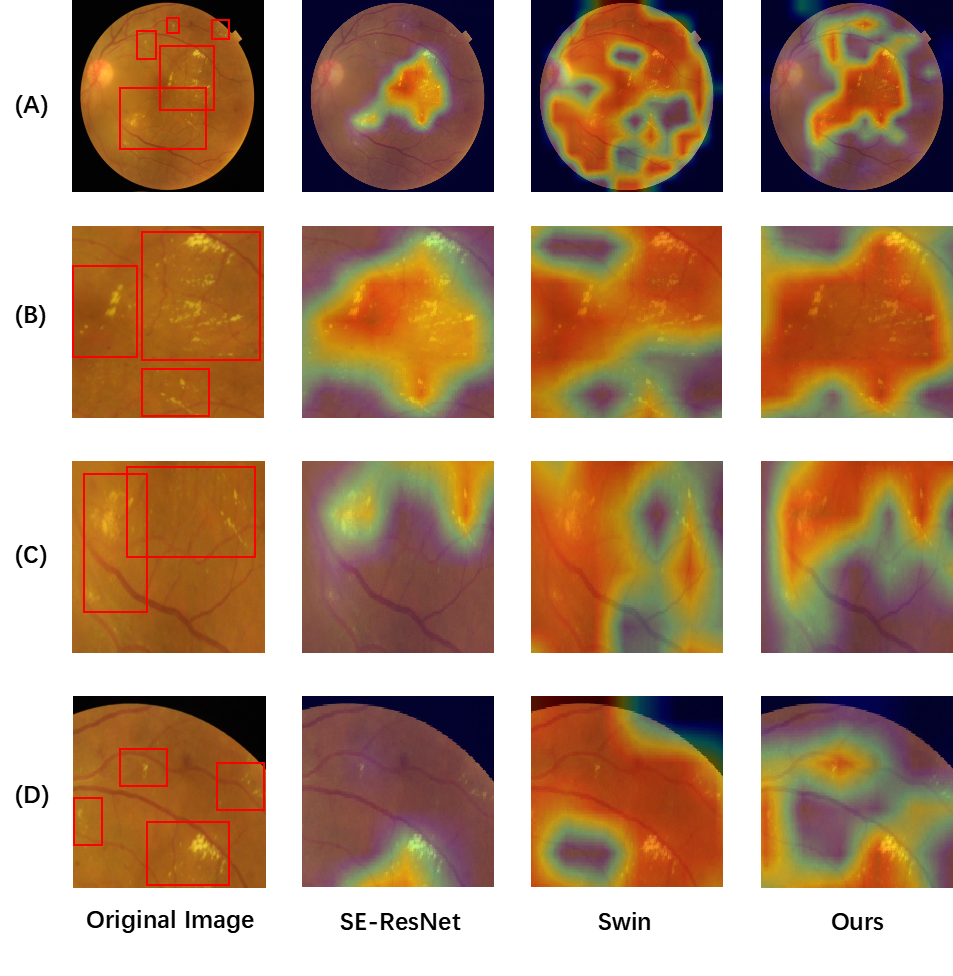}}
\caption{Comparison of Grad-CAM heatmaps of DR class obtained by different methods on RFMiD dataset. The lesion location is indicated by the red box. Row A displays the global view along with its corresponding heatmaps, while rows B, C and D represent local views and their respective heatmaps.}
\label{public}
\end{figure}

As shown in Table \ref{table_1}, our method achieve the best results in all four metrics on both the imbalanced dataset before resampling and the balanced dataset after resampling. Specifically, regarding the Accuracy score, our method reaches 80.29\% on the unbalanced OPTOS dataset, 2.7\% higher than Swin and 4.04\% higher than SE-ResNet \cite{hu2018squeeze}, which shows its ability to better capture the complex and subtle patterns in fundus images. To further strengthen our argument, we also test our method on the publicly available RFMiD dataset. Table \ref{table_2} shows that our method also achieves the highest scores in all metrics on balanced RFMiD.

To further analyze the classification results, we present confusion matrix results as shown in Fig. \ref{fig9}. Due to the similarity of features between Uveitis disease, Hypertension, and Normal on fundus images, it presents a challenge for models to classify these three types of diseases accurately. However, our approach outperforms SE-ResNet and Swin in the category of the three diseases. The result indicates our method can capture fine-grained features better.

\begin{figure}[ht]
\centerline{\includegraphics[width=\linewidth]{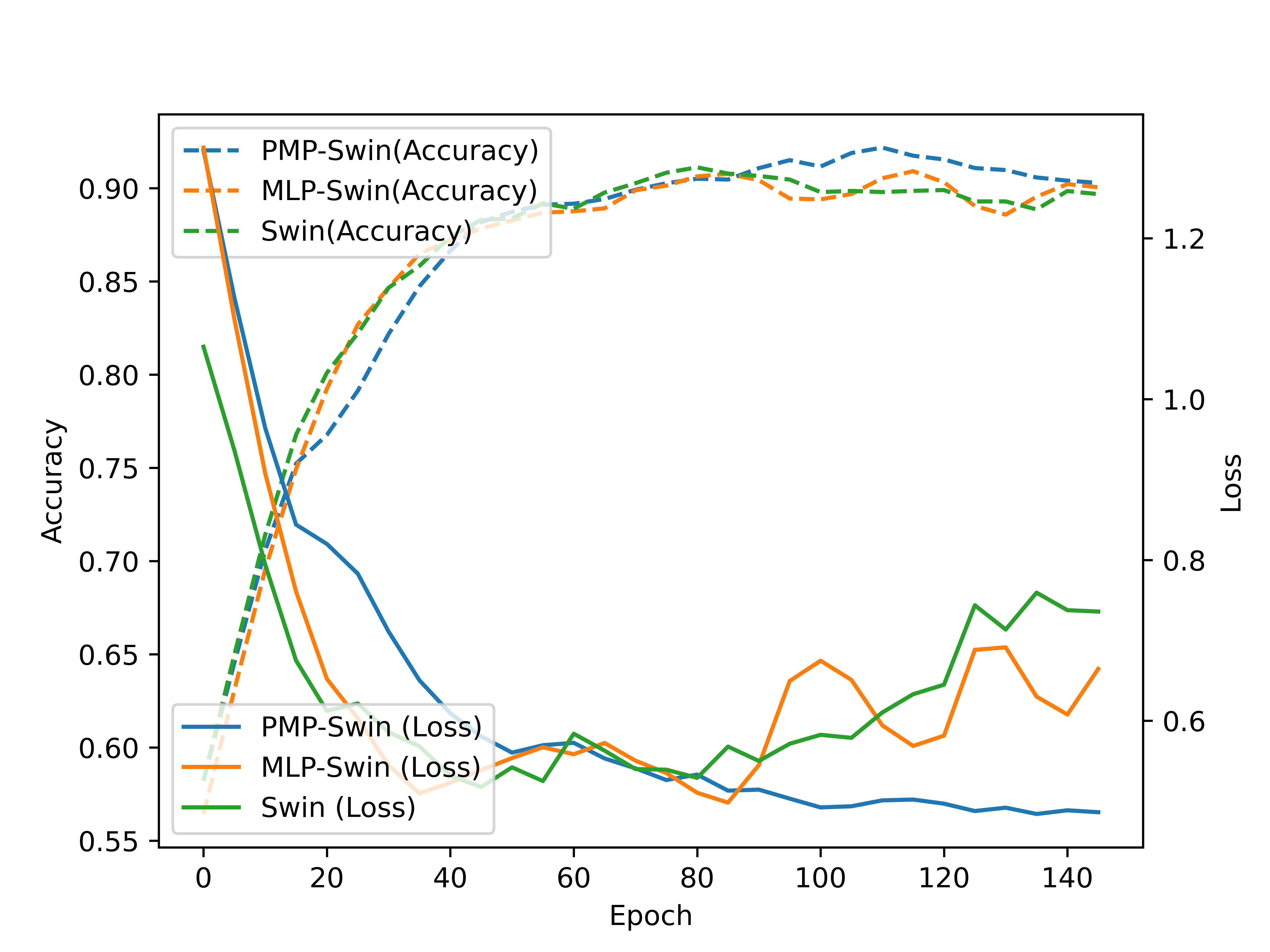}}
\caption{Comparison of valid accuracy and train loss curves obtained by methods based on Swin backbone on OPTOS dataset.}
\label{validation}
\end{figure}

In addition to the aforementioned quantitative result comparisons, we also use the Grad-CAM \cite{zhou2016learning} method to visualize the model performance. The color gradient indicates the degree of pixel attention in the heatmap generated by Pytorch Grad-CAM library \cite{jacobgilpytorchcam}, where deeper shades of red indicate greater attention and deeper shades of blue indicate weaker attention. It can be seen in Fig. \ref{heatmap} and Fig. \ref{public} that the focused area of SE-ResNet based on CNN structure is relatively concentrated due to its local receptive field. However, the focused areas of Swin and our method based on Transformer structure are dispersed due to their global receptive field, which can focus on more lesion areas. Compared to Swin, our method can more accurately focus on most lesion areas rather than the normal retinal background. In addition, as shown in Fig. \ref{public} for the DR class, our model can accurately identify hard exudates, as demonstrated by both the global heatmap and local heatmaps. 

\subsection{Ablation Studies}

This section presents our ablation study on the OPTOS dataset. Firstly, we validate the effectiveness of the PMP module and then analyze the robustness of our method with different parameters in our architecture design, including patch size, number of PMP modules, and the K value required for information aggregation.

\subsubsection{The effectiveness of Multi-Scale Patch Message Passing Module}

\begin{table*}[ht]
\caption{Comparison of Results Obtained By Different Combination Methods Based On Swin Structure On OPTOS Dataset}
\centering
\resizebox{\linewidth}{!}{
\begin{tabular}{c|ccc|llll|cccc}
\hline
\multirow{2}{*}{Method} &
  \multicolumn{3}{c|}{Combination} &
  \multicolumn{4}{c|}{Unbalanced} &
  \multicolumn{4}{c}{Balanced} \\ \cline{2-12} 
 &
  \multicolumn{1}{c|}{MLP} &
  \multicolumn{1}{c|}{PMP} &
  Multi-Scale &
  \multicolumn{1}{c|}{Accuracy(\%)} &
  \multicolumn{1}{c|}{Precision(\%)} &
  \multicolumn{1}{c|}{F1(\%)} &
\multicolumn{1}{c|}{Kappa(\%)} &
  \multicolumn{1}{c|}{Accuracy(\%)} &
  \multicolumn{1}{c|}{Precision(\%)} &
  \multicolumn{1}{c|}{F1(\%)} &
  Kappa(\%) \\ \hline
Swin &
  \multicolumn{1}{c|}{} &
  \multicolumn{1}{c|}{} &
   &
  \multicolumn{1}{l|}{77.59 $\pm$ 2.38} &
  \multicolumn{1}{l|}{75.20 $\pm$ 1.95} &
  \multicolumn{1}{l|}{75.37 $\pm$ 2.60} &
  73.19 $\pm$ 3.45&
  \multicolumn{1}{c|}{90.51 $\pm$ 3.19} &
  \multicolumn{1}{c|}{90.76 $\pm$ 2.30} &
  \multicolumn{1}{c|}{90.53 $\pm$ 2.96} &
  88.93 $\pm$ 4.36\\ \hline
MLP-Swin &
  \multicolumn{1}{c|}{\checkmark} &
  \multicolumn{1}{c|}{} &
   &
  \multicolumn{1}{l|}{75.48 $\pm$ 4.03} &
  \multicolumn{1}{l|}{74.51 $\pm$ 6.23} &
  \multicolumn{1}{l|}{73.16 $\pm$ 4.50} &
  70.66 $\pm$ 5.42&
  \multicolumn{1}{c|}{90.34 $\pm$ 0.92 } &
  \multicolumn{1}{c|}{90.50 $\pm$ 0.75} &
  \multicolumn{1}{c|}{90.33 $\pm$ 0.87} &
  88.73 $\pm$ 1.25\\ \hline
PMP(Mono)-Swin &
  \multicolumn{1}{c|}{} &
  \multicolumn{1}{c|}{\checkmark} &
   &
  \multicolumn{1}{l|}{76.54 $\pm$ 3.19} &
  \multicolumn{1}{l|}{75.04 $\pm$ 3.18} &
  \multicolumn{1}{l|}{74.21 $\pm$ 1.90} &
  71.91 $\pm$ 3.66 &
  \multicolumn{1}{c|}{91.94 $\pm$ 0.30} &
  \multicolumn{1}{c|}{\textbf{92.39 $\pm$ 0.31}} &
  \multicolumn{1}{c|}{92.03 $\pm$ 0.29} &
  90.60 $\pm$ 0.41\\ \hline
\textbf{PMP-Swin(Ours)} &
  \multicolumn{1}{c|}{} &
  \multicolumn{1}{c|}{\checkmark} &
  \checkmark &
  \multicolumn{1}{l|}{\textbf{80.29 $\pm$ 1.04}} &
  \multicolumn{1}{l|}{\textbf{78.60 $\pm$ 0.75}} &
  \multicolumn{1}{l|}{\textbf{78.00 $\pm$ 0.79}} &
  \textbf{76.36 $\pm$ 1.38} &
  \multicolumn{1}{c|}{\textbf{92.12 $\pm$ 0.27}} &
  \multicolumn{1}{c|}{92.29 $\pm$ 0.36} &
  \multicolumn{1}{c|}{\textbf{92.13 $\pm$ 0.29}} &
  \textbf{90.80 $\pm$ 0.37}\\ \hline
\end{tabular}
}
\label{table_3}
\end{table*}

Table \ref{table_3} demonstrates that the introduction of the PMP module can effectively improve the performance of the Swin backbone. To be more specific, the Accuracy score of PMP-Swin increases by 2.7\% on the unbalanced OPTOS dataset and 1.61\% on the balanced OPTOS dataset. However, replacing PMP modules with MLP layers results in decreased performance compared with Swin, especially on the unbalanced OPTOS dataset, with the Accuracy score decreasing by 2.11\%. It may be because constructing more complex feature representations of semantic features with simple linear layers cannot exploit useful information for better classification. 

To verify the necessity of multi-scale branches, we compare the performance of PMP(Mono)-Swin with only one branch of PMP modules, and PMP-Swin with two branches of PMP modules. As shown in Table \ref{table_3}, the classification performance of PMP-Swin is better than that of PMP(Mono)-Swin both before and after resampling, indicating that multi-scale branches can effectively improve classification performance. This may be due to that the diversity of lesion area sizes is taken into account when aggregating information in patches input to PMP modules, which is beneficial for establishing global connections of lesion features. Additionally, training of methods based on Swin backbone with different structures is given in Fig. \ref{validation}. It can be observed that Swin and MLP-Swin overfit after training for 80 epochs, while our method's loss curve continues to decrease until around 140 epochs, implying that our method can effectively prevent overfitting and achieve higher classification accuracy.

\subsubsection{Comparison of different structure settings}

\begin{table}[ht]
\caption{Comparison of Results Obtained By Different Structure Settings of PMP-Swin On OPTOS Dataset}
\centering
\resizebox{\linewidth}{!}{
\begin{tabular}{c|ccc|ccc|c}
\hline
\multirow{2}{*}{Method} & \multicolumn{3}{c|}{Small Branch}                   & \multicolumn{3}{c|}{Large Branch}                    & \multirow{2}{*}{Performance(\%)} \\ \cline{2-7}
 &
  \multicolumn{1}{c|}{\begin{tabular}[c]{@{}c@{}}Patch \\ Size\end{tabular}} &
  \multicolumn{1}{c|}{K} &
  N &
  \multicolumn{1}{c|}{\begin{tabular}[c]{@{}c@{}}Patch\\ Size\end{tabular}} &
  \multicolumn{1}{c|}{K} &
  N &
   \\ \hline
A                       & \multicolumn{1}{c|}{1} & \multicolumn{1}{c|}{8} & 3 & \multicolumn{1}{c|}{4} & \multicolumn{1}{c|}{2} & 3 & 91.82 $\pm$ 0.72                         \\ \hline
B                       & \multicolumn{1}{c|}{2} & \multicolumn{1}{c|}{8} & 3 & \multicolumn{1}{c|}{4} & \multicolumn{1}{c|}{2} & 3 & 91.50 $\pm$ 0.58                       \\ \hline
C                       & \multicolumn{1}{c|}{1} & \multicolumn{1}{c|}{4} & 3 & \multicolumn{1}{c|}{4} & \multicolumn{1}{c|}{2} & 3 & 91.23 $\pm$ 0.50                       \\ \hline
D                       & \multicolumn{1}{c|}{1} & \multicolumn{1}{c|}{8} & 2 & \multicolumn{1}{c|}{4} & \multicolumn{1}{c|}{2} & 3 & 91.22 $\pm$ 0.63                        \\ \hline
E                       & \multicolumn{1}{c|}{1} & \multicolumn{1}{c|}{8} & 3 & \multicolumn{1}{c|}{8} & \multicolumn{1}{c|}{2} & 3 & 91.79 $\pm$ 0.55                       \\ \hline
F                       & \multicolumn{1}{c|}{1} & \multicolumn{1}{c|}{8} & 3 & \multicolumn{1}{c|}{4} & \multicolumn{1}{c|}{2} & 2 & 91.25 $\pm$ 0.82                        \\ \hline
\textbf{PMP-Swin(Ours)}                       & \multicolumn{1}{c|}{1} & \multicolumn{1}{c|}{8} & 3 & \multicolumn{1}{c|}{4} & \multicolumn{1}{c|}{4} & 3 & \textbf{91.84 $\pm$ 0.31}              \\ \hline
\end{tabular}
}
\label{table_5}
\end{table}

Table \ref{table_5} shows the impact of different parameter settings on the average performance of our method across four metrics. 
With different settings, all models reach a high accuracy of over 91\%, supporting our method's effectiveness for robustness. Firstly, for patch size, we select different pairs of patch sizes including (1,4), (2,4), and (1,8), and the results showed that (1,4) achieved the best average performance. K represents how many most similar patches in the feature space each patch needs to select for information aggregation. Our experiments show that when the K of the small branch is 8 and the K of the large branch is 4, our method achieves the best performance. This issue may be attributed to the size of the patch. To be more specific, patch size determines the total number of patches. Secondly, when the patch size is too small, K should be appropriately increased, otherwise, it will lead to information aggregation only in local regions. When the patch size is too large, K should be reduced, otherwise, it will lead to information aggregation between patches containing semantic information and those containing too much non-lesion semantic information. In addition, we also consider the impact of the depth of both branches on the model. The best performance is achieved when both branches have a depth of 3. It can be seen that when N is reduced, the model's performance declines, indicating that a sufficient stack of PMP modules can extract more features.

\subsection{CONCLUSION}

In this paper, we present a novel framework named PMP-Swin for the classification of retinal diseases in fundus images, which incorporates two key improvements. Firstly, we introduce a new PMP module based on the Message Passing mechanism that can be easily integrated into the Transformer Backbone. This enables us to fully exploit patch features and establish global connections among disease-related features. Secondly, we utilize a dual-branch approach with PMP modules to learn features at multiple scales considering that the scale of pathological features varies among different diseases. Our extensive experiments on both private and public datasets demonstrate that our method outperforms current state-of-the-art techniques based on CNNs and Transformers. We believe that our proposed method can inspire more Transformer-based classification diagnostic techniques, which will further promote the application of deep learning in clinical diagnosis.

\bibliographystyle{IEEEtran}
\bibliography{references}
\end{document}